\documentclass{article}

\usepackage{arxiv}

\usepackage[utf8]{inputenc} 
\usepackage[T1]{fontenc}    
\usepackage{hyperref}       
\usepackage{url}            
\usepackage{booktabs}       
\usepackage{amsfonts}       
\usepackage{nicefrac}       
\usepackage{microtype}      
\usepackage{lipsum}		
\usepackage{graphicx}
\usepackage[numbers,square]{natbib}
\usepackage{amsmath,amssymb,amsfonts}
\usepackage{doi}

\title{An End-to-End Point of Interest (POI) Conflation Framework}

\author{ 

    \href{https://orcid.org/0000-0003-3885-0812}{\includegraphics[scale=0.06]{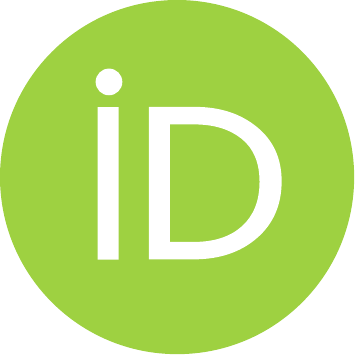}\hspace{1mm}Raymond Low} \\
	Engineering Systems and Design Pillar\\
	Singapore University of Technology and Design\\
	8 Somapah Rd, Singapore 487372 \\
	\texttt{iamraymondlow@gmail.com} \\
	
	\And
	\href{https://orcid.org/0000-0002-1858-0846}{\includegraphics[scale=0.06]{orcid.pdf}\hspace{1mm}Zeynep D. Tekler} \\
	Engineering Product Development Pillar\\
	Singapore University of Technology and Design\\
	8 Somapah Rd, Singapore 487372 \\
	\texttt{duygutekler.zeynep@gmail.com} \\
	
	\And
	\href{https://orcid.org/0000-0001-6312-0331}{\includegraphics[scale=0.06]{orcid.pdf}\hspace{1mm}Lynette Cheah} \\
	Engineering Systems and Design Pillar\\
	Singapore University of Technology and Design\\
	8 Somapah Rd, Singapore 487372 \\
	\texttt{lynette@sutd.edu.sg} \\
}

\renewcommand{\shorttitle}{An End-to-end POI Conflation Framework}

\hypersetup{
pdftitle={A template for the arxiv style},
pdfsubject={q-bio.NC, q-bio.QM},
pdfauthor={David S.~Hippocampus, Elias D.~Striatum},
pdfkeywords={First keyword, Second keyword, More},
}

\begin{document}
\maketitle

\begin{abstract}
Point of interest (POI) data serves as a valuable source of semantic information for places of interest and has many geospatial applications in real estate, transportation, and urban planning. With the availability of different data sources, POI conflation serves as a valuable technique for enriching data quality and coverage by merging the POI data from multiple sources. This study proposes a novel end-to-end POI conflation framework consisting of six steps, starting with data procurement, schema standardisation, taxonomy mapping, POI matching, POI unification, and data verification. The feasibility of the proposed framework was demonstrated in a case study conducted in the eastern region of Singapore, where the POI data from five data sources was conflated to form a unified POI dataset. Based on the evaluation conducted, the resulting unified dataset was found to be more comprehensive and complete than any of the five POI data sources alone. Furthermore, the proposed approach for identifying POI matches between different data sources outperformed all baseline approaches with a matching accuracy of 97.6\% with an average run time below 3 minutes when matching over 12,000 POIs to result in 8,699 unique POIs, thereby demonstrating the framework's scalability for large scale implementation in dense urban contexts.
\end{abstract}


\section{Introduction}
The ubiquitous use of mobile devices, combined with advancements in location-aware technologies, has increased our ability to capture individual mobility data at increasing geospatial-temporal resolutions \cite{miller2015geographic, tekler2020scalable}. A potential application of this capability includes the identification of points of interest (POI) by analysing users' mobility data to identify specific locations of interest that are regularly visited by the same user or by a large number of distinct users throughout the day \cite{guidotti2014retrieving, vhaduri2017discovering}. Other than passively analysing the users' mobility data, a more active data collection approach involves crowdsourcing, where a community of volunteers are asked to provide semantic information about a recently visited location to assemble and maintain a high-resolution geospatial database \cite{touya2017assessing}.  Some of the semantic information includes the location's name, address, opening hours, geographic coordinates, user reviewers, and place type categorisation. With the continual changes in POI data over time due to business renewals and urban development, this valuable source of geospatial data continues to remain relevant with many potential application areas in various fields.
 
\subsection{Applications of POI Data}
Past studies have applied the use of POI data in a wide range of application areas. For instance, Gong et al. \cite{gong2016inferring} combined the use of taxi trajectory data with POI data to infer the passengers' trip purpose at each stop, while Liu et al. \cite{liu2020identification} used a similar dataset to perform land use classification of different urban regions in Chengdu, China. Within the context of urban transportation, Low et al. \cite{low2020commercial} also combined the use of POI data with trip-related information obtained from a commercial vehicle travel survey to infer the activities conducted at each stop during a vehicle tour. POI data has also been utilised in urban studies where researchers have attempted to perform disaggregated employment estimation based on the POIs found within a particular region \cite{rodrigues2013estimating}. Given the number of application areas that could potentially benefit from the availability of POI data, there is, fortunately, multiple data sources for obtaining this valuable geospatial information.

\subsection{POI Data Sources}
These data sources can be broadly grouped into one of four categories: open-sourced projects, commercial data providers, government agencies, and location-based social networks (LBSN).

Open-sourced projects are typically supported by a community of volunteers that contribute to an open-sourced geospatial database through a crowdsourcing approach. More specifically, some of these contributions involve updating the semantic information of existing POIs, removing outdated POIs, adding new POIs, and validating any potential changes to the database proposed by other contributors. The latest version of the dataset is also typically released to the general public to be used freely at no monetary cost. Open Street Map (OSM) is a prime example of such a project that has been conducted on a global scale.

Commercial data providers and government agencies, on the other hand, maintain a database of different business establishments and critical facilities to support various applications, including commercial market research, policymaking \cite{tekler2019waste}, and urban planning. While the POI data by these two sources are often not accessible to the general public due to proprietary reasons, private access to the database can sometimes be obtained by leasing it from the respective government agencies and data providers through an annual subscription plan.

Lastly, LBSNs rely on their vast network of end-users to maintain the relevancy of their database by encouraging their users to share their location information and visiting experiences with other users on the platform in the form of user reviewers and ratings. Some platforms even rely on the users' smartphone connection to nearby cell towers and wireless networks to infer the users' last visited locations within the building by combining it with various indoor localisation techniques \cite{tekler2019alternative,farshad2013microscopic}. Some examples of these LBSNs includes Swarm by Foursquare \cite{swarm2020} and Google Maps \cite{googlemap2020}. 

Table \ref{tab:POIsource} provides a list of POI data sources grouped based on the four categories described above.

\begin{table}
\centering
\caption{Examples of POI data sources grouped based on four different categories: open-sourced projects, commercial data providers, government agencies, and LBSNs. Each data source is tagged correspondingly to indicate whether it provides global or regional coverage.}
\label{tab:POIsource}
\setlength{\tabcolsep}{2pt}

\begin{tabular}{ll}
\hline
{Categories} & {POI Sources} \\ \hline

{\begin{tabular}[c]{@{}l@{}}Open-source \\ projects\end{tabular}} & {\begin{tabular}[c]{@{}l@{}} OpenStreetMap (Global) \cite{osm2020}, GeoNames (Global) \cite{geonames2020}\end{tabular}} \\ \hline

{\begin{tabular}[c]{@{}l@{}}Commercial \\ providers\end{tabular}} &
  {\begin{tabular}[c]{@{}l@{}} Dun \&   Bradstreet \cite{dunbradstreet2020} (Global), InfoUSA (Regional) \cite{infousa2020}\end{tabular}} \\ \hline
  
{\begin{tabular}[c]{@{}l@{}}Government \\ agencies \end{tabular}} &
  {\begin{tabular}[c]{@{}l@{}} OneMap (Regional) \cite{onemap2020}\end{tabular}} \\ \hline
  
{\begin{tabular}[c]{@{}l@{}}Location- \\ based social \\ networks\end{tabular}} &
  {\begin{tabular}[c]{@{}l@{}}Google Places (Global) \cite{googleplacetypes2021}, HERE Map (Global) \cite{here2020}, \\ Foursquare (Global) \cite{foursquare2020}, Yelp (Global) \cite{yelp2020}, \\ Baidu Map (Global) \cite{baidu2020}, Weibo (Regional) \cite{weibo2020}, \\ Facebook (Global) \cite{facebook2020}, Yahoo! (Global) \cite{yahoo2020}, \\ Trip Advisor (Global) \cite{tripadvisor2020}, Gaode Map (Regional) \cite{gaode2020}\end{tabular}} \\ \hline
\end{tabular}
\end{table}

\subsection{POI Conflation}
With a large number of POI data sources available to choose from, there are many potential benefits of conflating multiple POI sources to obtain a single unified dataset. These benefits include (i) the ability to combine the complementary attributes found in different data sources to enrich the semantic information stored in each POI, (ii) increasing the data coverage and richness of the resulting dataset, and (iii) improving the resulting data quality by correcting for any erroneous or missing information.

However, there are many technical challenges that need to be addressed when performing POI conflation. The first challenge is related to different data sources using non-standardised schemas or data formats when storing the attributes of their POIs. A typical example is the use of different attribute names when referring to the same attribute (i.e., place type, location category, location type, venue category). This issue can result in complications during the POI matching step, where we attempt to identify overlapping POIs between different data sources by comparing their POI attributes. Another challenge encountered during POI conflation involves standardising the diverse taxonomies used by different data sources when categorising the function of the same POI. For instance, a POI categorised as a "restaurant" in one data source can also be categorised as "eatery" in another data source. Lastly, it is also crucial to ensure that the POI matching process is computationally efficient to maintain its viability when applied over an extensive geographical area of interest involving a large number of POIs. Many of these challenges increase exponentially when many POI sources are required to be conflated simultaneously.

\subsection{Study Objective and Contributions}
This paper proposes a novel framework for performing end-to-end POI conflation involving a six-step approach. The framework begins with the data procurement step, which involved gathering POI data from various data sources before formatting the data to follow a custom schema in the schema standardisation step. Due to the distinct place type taxonomies adopted by each data source, a taxonomy mapping step is subsequently performed to ensure that all POI data follow a standard taxonomy. Once all POIs are formatted based on the same custom schema while following a consistent place type taxonomy, the POI matching step is performed to identify any overlapping POIs among the different data sources. The matching POIs identified are conflated in the POI unification step, and the resulting unified dataset was verified in the final data verification step. The feasibility of the proposed framework was demonstrated in a case study conducted within Singapore, where the POI data from five different data sources was simultaneously conflated to form a unified POI dataset. This work contributes to the literature as a more comprehensive and end-to-end POI conflation framework that has been evaluated based on real-world geospatial datasets and is viable for large-scale implementations.

\section{Literature Review}
\label{sec:literaturereview}
This section provides a thorough review of the existing literature related to POI matching and POI conflation, where the former is an essential step performed during POI conflation.

\subsection{POI matching}
\label{sec:poimatching}
POI matching refers to the process of identifying matching POIs between different data sources based on the similarity in their semantic attributes, including geospatial coordinates, location name, address, place type, and description. Therefore, POI matching can be viewed as an extension of toponym matching, which mainly involves the identification of matching geographical locations by comparing the character strings in their location names \cite{santos2018learning, santos2018toponym, kilincc2016accurate}.

A study conducted by McKenzie et al. \cite{mckenzie2014weighted} used a weighted combination of the location name, geographic distance, and topic similarity metrics to identify POI matches in Yelp and Foursquare. A binomial probit regression model was used to estimate the overall contribution of each attribute, resulting in a matching accuracy of 97\% for 100 randomly selected POIs. An entropy-weighted approach was also introduced by Li et al. \cite{li2016entropy} that uses spatial, name, and place type similarity measures to identify POI matches between Baidu Map and Sina. In their study, word segmentation and phonetic-based methods were adopted to avoid any semantic ambiguity, and a mapping between different place type taxonomies was performed to address the issues of heterogeneity and semantic relatedness, resulting in a final f1-score of 0.85. However, it should be noted that the proposed taxonomy mapping approach is designed explicitly for taxonomies that follow a hierarchical tree structure. Lastly, a study conducted by Li et al. \cite{li2020different} proposed a POI matching approach that first performs a multi-attribute constraint calculation of the name, address, class, and spatial similarity metrics, before manually determining the thresholds of these constraints based on their f1-scores. This approach was tested on POI data from Baidu Map and Gaode Map to result in a final f1-score of 96.9\% in the test area.

Other than adopting a weighted multi-attribute matching approach, several studies have also proposed other algorithms to aggregate various similarity measures for POI matching. Novack et al. \cite{novack2018graph} proposed a graph-based matching approach to match the POIs from two different data sources (i.e., Foursquare and OSM) by representing each POI as a node in a graph and using the edges to represent the matching possibilities between each POI. The evaluation between each matching pair was based on three similarity measures, including spatial, name, and semantics similarity. By using a simple weighted approach to aggregate these similarity measures, three different graph-based matching strategies (i.e., Naive Matching, Best-best Matching, and Combinatorial Matching) were proposed and evaluated on a test area in London to result in an overall matching accuracy of 86\%. While the authors claimed that the approach is scalable when applied to larger areas, the claim may not hold when conflating multiple POI sources as it will increase the number of potential edges that can be formed between each node. Another study conducted by Psaila and Toccu \cite{psaila2019fuzzy} proposed an approach based on fuzzy logic and possibility theory to perform online aggregation of POIs from Google Places and Facebook. The proposed approach measures the degree of likelihood between two place descriptors, containing information about the location name, address, and geographic coordinates, to evaluate if they refer to the same location. The approach's effectiveness was tested in three cities, Manchester, Genoa, and Stuttgart, and reported f1-scores of up to 93.1\%. Another study conducted by Yu et al. \cite{yu2018holistic} proposed a framework to aggregate several similarity metrics through approval voting to perform POI matching between OSM and the GeoNames gazette without any parameter tuning. The similarity metrics considered in this study include spatial, name, structural, and extensional similarity. Another related study was conducted by Almedia et al. \cite{almeida2018automatic}, who proposed a POI matching approach based on an outlier detection model. The study began by identifying POI matches using the Factual Crosswalk API to connect the POIs from the Factual database with their Facebook and Foursquare counterparts before using them to train a machine learning (ML) model to perform outlier detection. By testing out different combinations of string comparison approaches for the name, website, address, and category attributes, the best model resulted in a matching accuracy of 94.7\% and a ROC score of 0.975. Lastly, a study conducted by Jiang et al. \cite{jiang2015mining} proposed a method using the JaroWinklerTFIDF algorithm \cite{cohen2003comparison} to standardise the place type taxonomy used in Yahoo! to follow the North American Industry Classification System (NAICS) before performing POI matching between Yahoo! and several proprietary datasets. By identifying matches with high similarity scores, these matches were subsequently used as training data to develop the ML models needed to perform POI classification for matches with a lower similarity score. 

\subsection{Past Works on POI Conflation}
On the other hand, significantly fewer studies have explored the topic of POI conflation as it requires a further investigation on the other steps, such as the unification process, after identifying the matching POIs through POI matching.

A study conducted by Yang et al. \cite{yang2015pattern} proposed a novel pattern-mining approach for conflating road networks with POI data. The proposed approach involves generating and aligning the pattern-related skeleton graphs for the POIs and road networks before comparing the semantic data from the two data sources to infer the road names of the road segments. Another study conducted by Yu et al.  \cite{yu2018semantic} attempted to automate the geospatial data conflation process by first transforming different data sources to a designated ontology before using a series of semantic web rule language (SWRL) rules to find matching POIs and resolve any conflicts during the conflation process.

By comparing against the studies reviewed in this section, the novel POI conflation framework proposed in this study stands as a more comprehensive end-to-end approach, starting with the data procurement process and ending with a data verification step after identifying and unifying the matching POIs from different data sources. Furthermore, to ensure that the framework is generalisable to a wide range of data sources containing different sets of POI attributes, the framework was also successfully applied on five real-world POI datasets in a case study conducted in Singapore.

\section{POI Conflation Framework: Overview}
\label{sec:framework}
This section provides an overview of the proposed POI conflation framework, which consists of six steps:

\begin{enumerate}
  \item \textbf{Data procurement}: The data procurement step involves the process of extracting, gathering, or downloading POI data from various data sources in their original data format and schemas for the study area of interest.
  
  \item \textbf{Schema standardisation}: After procuring the POI data from their respective sources, the schema standardisation step is performed to standardise the storage format of the POIs obtained based on a custom schema.
 
  \item \textbf{Taxonomy mapping}: Due to the unique taxonomies adopted by different data sources when categorising their POI data, a taxonomy mapping step is performed to standardise the categorisation or classification of each POI based on a singular taxonomy.
  
  \item \textbf{POI matching}: Once all POIs are formatted based on the same custom schema while following a consistent place type taxonomy, the POI matching step involves identifying the overlapping POIs between different data sources by comparing the similarities between their semantic attributes (i.e., geospatial coordinates, location name, address, place type and description).

  \item \textbf{POI unification}: After identifying the matching POIs between different data sources, the POI unification step involves combining the semantic attributes of the matching POIs while improving the data quality of the resulting dataset by correcting for any erroneous information or missing fields.

  \item \textbf{Data verification}: The final data verification step is performed to verify the conflated POI dataset either manually through the employment of human domain experts or programmatically using established data validation metrics.
 
\end{enumerate}

\begin{figure}[t]
\centering
\includegraphics[width=0.75\textwidth]{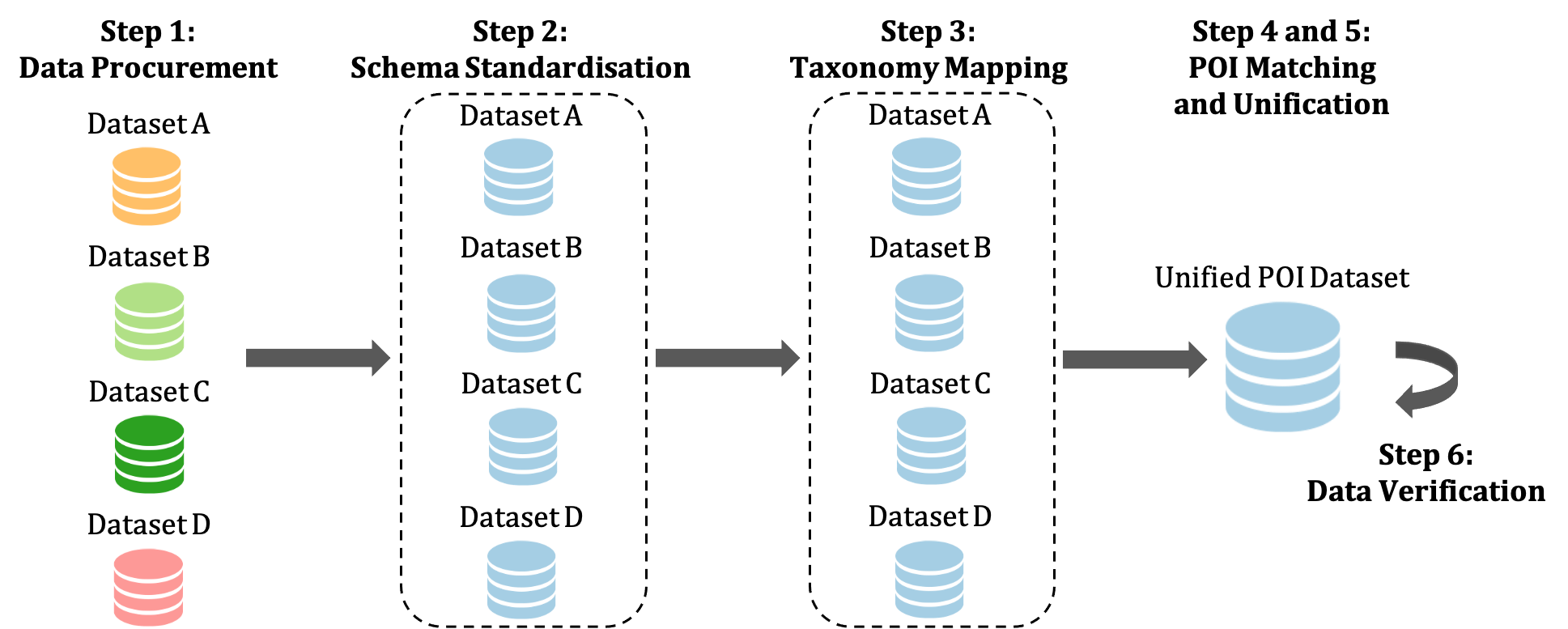}
\caption{Overview of the proposed POI conflation framework. \label{fig:overview}}
\end{figure}

A graphical representation of the proposed POI conflation framework is provided in Figure \ref{fig:overview}.

\section{Case Study}
A case study is conducted in a study area within the island state of Singapore involving five POI data sources to demonstrate the feasibility of the proposed POI conflation framework. It should be noted that while the framework was applied to a specific study area as part of this work, the steps described can be easily replicated in other geographical locations and on other POI data sources.

\subsection{Study Area}
The study area chosen for this case study is the residential town of Tampines (Southwest: 1.310157, 103.923457; Northwest: 1.374323, 103.923457; Northeast: 1.374323, 103.987876; Southeast: 1.310157, 103.987876), which is located in the eastern region of Singapore. Tampines is the third-largest town in the island state, with a geographic area spanning over 20.9 km\textsuperscript{2} and housing a total population of 237,800 in 2018 \cite{datagovsg_master_2021, landarea2018}. A wide diversity of amenities can also be found in the study area, including public transit nodes, community centres, retail malls, schools, and healthcare facilities, along with residential areas and business parks, hosting a multitude of industrial estates. Given the multitude of amenities and land-use types found within the study area, a diverse and comprehensive range of POI data can be found within the study area. On top of that, due to the local government's continued efforts towards data sharing through their Open Data initiatives \cite{datagovsg2021}, this allows us to easily access POI data from local government agencies, on top of those obtained from open-sourced projects, commercial data providers, and LBSNs, for this study.

\subsection{Data Description}
This section provides a thorough description of the five POI data sources considered for this study: OpenStreetMap (OSM), Google Places, HERE Map, OneMap, and the Singapore Land Authority (SLA) 2020 dataset. The first three data sources were chosen due to their prevalent use in the literature and coverage within the study area, while the last two sources were selected to represent data from the government agencies.
 
\subsubsection{OpenStreetMap (OSM)}
OSM is a prime example of an open-source project that relies on a community of volunteers to develop and maintain a public geospatial database on a global scale through a crowdsourcing approach. Full access to the OSM database has been made freely available online due to the initiative's dedication to encouraging the growth, development, and distribution of free geospatial data. Users are provided with various options to download the dataset in bulk at different geographic scales (i.e., planet, continent, country, and metropolitan area) or extract the POI data from specific regions via the Overpass API \cite{osmoverpass2020}. On top of that, the database's update frequency ranges from a weekly basis for the entire planet down to a minute-by-minute real-time update depending on specific regions and countries \cite{osmplanet2020}. Despite the easy accessibility of the database, the heavy reliance on a crowdsourcing approach for data procurement and maintenance has led to issues related to data inconsistencies \cite{haklay_how_2010-1} and the presence of incomplete entries due to differing standards amongst the contributors. These factors negatively impact the dataset's data quality and limit its use in various geospatial applications.
 
\subsubsection{Google Places}
Google Places is a web mapping platform developed by Google, providing end-users with different mapping services such as real-time updates on traffic conditions, route planning for different travel modes, satellite imagery, and panoramic street views. The platform relies on a range of approaches such as satellite imagery, authoritative sources (e.g., local government agencies, non-government organisations, private data providers), and timely feedback from existing platform end-users to maintain the relevancy of its geospatial database. Therefore, this data source falls into the category of an LBSN. Until recently, the organisation has also begun leveraging on the advancements in ML to automate and improve the accuracy of the mapping process by using computer vision to identify the outlines of road networks and buildings \cite{google1012020}. While the POI data from Google Places cannot be downloaded in bulk, unlike in OSM, users who are interested in leveraging this comprehensive database can obtain detailed POI information about a specific geographical location by using the Places API \cite{googleplacetypes2021} at a small cost.

\subsubsection{HERE Map}
HERE Map is an example of a commercial data provider that provides customers with a rich set of geospatial data to support their mapping needs. While the company advertises the use of state-of-art technology and leading mapping processes to assemble and maintain its geospatial database \cite{heredata2020}, the exact details of these processes cannot be found in their online documentation and are assumed to be proprietary. Users of their service can obtain POI data for a particular region either by using the HERE RESTful API, subjected to monthly transaction limits \cite{hereapi2020}, or leased in bulk through a data subscription plan. Users can also report any map inconsistencies by utilising the Map Feedback API \cite{herefeedback2020} provided by the platform. 

\subsubsection{OneMap}
OneMap is the authoritative national map of Singapore that was developed by the Singapore Land Authority (SLA). The mapping platform was created with the objective of providing location-based services to its end-users through the support of various government agencies. Some of these services include providing (i) bus arrival timings and route information, (ii) land use and ownership information, (iii) locations of nearby educational institutes, as well as (iv) traffic conditions and parking availability \cite{onemap2020}. Users of the mapping service can also utilise the OneMap RESTful API to query for different POIs within the country based on their thematic information, including parking lots, hospitals, restaurants, national parks, historical sites, museums, and transit nodes \cite{onemapthemes2021}.

\subsubsection{SLA 2020 Dataset}
 The SLA 2020 dataset is another geospatial dataset maintained by SLA to guide future governance policies related to land development, housing allocation, critical infrastructure, and transportation planning. This dataset differs from the OneMap dataset as it can only be obtained by directly licensing it from SLA on an annual basis and is not readily accessible to the general public due to the data's sensitivity. Apart from the location name and address information, each POI in the dataset is categorised based on 55 different place types, including education institutions, transportation ports, religious buildings, local government offices and critical healthcare facilities.
 
 Table \ref{tab:POIsummary} provides a summary of the five POI data sources considered in this study, covering information about how their data is procured and validated, as well as their update frequencies, limitations and place type coverage.

\begin{table*}
\centering
\caption{Summary of the five POI data sources considered in this study, covering information about how their data is procured and validated, as well as their update frequencies, limitations and place type coverage \cite{osmmapfeatures2021, googleplacetypes2021, hereplacetypes2021, onemapthemes2021}.}
\label{tab:POIsummary}
\setlength{\tabcolsep}{3pt}
\begin{tabular}{lllllll}
\textbf{Source} &
  \textbf{Procurement} &
  \textbf{Validation} &
  \textbf{Extraction} &
  \textbf{Update} &
  \textbf{Categories} &
  \textbf{Remarks} \\ \hline
OSM &
  Crowdsourcing &
  Crowdsourcing &
  \begin{tabular}[c]{@{}l@{}}Bulk online \\ download or\\ Overpass API\end{tabular} &
  \begin{tabular}[c]{@{}l@{}}Up to \\ weekly\\ basis\end{tabular} &
  \begin{tabular}[c]{@{}l@{}}29 main\\ and > 87\\ subcategories\end{tabular} &
  Data inconsistency \\ \hline
\begin{tabular}[c]{@{}l@{}}Google\\ Places\end{tabular} &
  \begin{tabular}[c]{@{}l@{}}Satellite imagery,\\ authoritative\\ bodies, and\\ crowdsourcing\end{tabular} &
  \begin{tabular}[c]{@{}l@{}}Data validation\\ team and\\ crowdsourcing\end{tabular} &
  Places API &
  Unclear &
  \begin{tabular}[c]{@{}l@{}}> 96 place\\ types\end{tabular} &
  \begin{tabular}[c]{@{}l@{}}Commercial\\ service\end{tabular} \\ \hline
\begin{tabular}[c]{@{}l@{}}HERE\\ Map\end{tabular} &
  Unclear &
  \begin{tabular}[c]{@{}l@{}}Some\\ crowdsourcing\end{tabular} &
  \begin{tabular}[c]{@{}l@{}}Places \\ (Search) API\end{tabular} &
  Unclear &
  \begin{tabular}[c]{@{}l@{}}> 166 place\\ types\end{tabular} &
  \begin{tabular}[c]{@{}l@{}}Commercial \\ service subjected \\ to transaction limits\end{tabular} \\ \hline
OneMap &
  \begin{tabular}[c]{@{}l@{}}Government\\ agencies\end{tabular} &
  \begin{tabular}[c]{@{}l@{}}Government\\ agencies\end{tabular} &
  OneMap API &
  Unclear &
  > 63 themes &
  None \\ \hline
\begin{tabular}[c]{@{}l@{}}SLA\\ Dataset\end{tabular} &
  \begin{tabular}[c]{@{}l@{}}Government\\ agencies\end{tabular} &
  \begin{tabular}[c]{@{}l@{}}Government\\ agencies\end{tabular} &
  Licensing &
  Yearly &
  \begin{tabular}[c]{@{}l@{}}55 place \\ types\end{tabular} &
  \begin{tabular}[c]{@{}l@{}}Not accessible to\\ general public\end{tabular}
\end{tabular}
\end{table*}

\subsection{Application within study area}

\subsubsection{Step 1: Data procurement}
The framework begins with the data procurement step, which involved gathering POI data from the five data sources (i.e., OSM, Google Places, HERE Map, OneMap, and SLA 2020 Dataset) in July-August 2021.

The data procurement process for OSM and the SLA dataset is relatively straightforward as the POIs in the study area can be downloaded in their entirety through the OSM website or licensed directly from the appropriate government agency. On the other hand, the POI data for the remaining sources (i.e., OneMap, Google Places, and HERE Map) can only be obtained through their respective APIs. Each API call is constructed by providing a unique API key for authentication purposes and allows users to provide additional parameters to refine the query. For instance, users of OneMap are required to provide the themes of the POIs that they are interested in querying within the query string, which is equivalent to the place type attribute found in other data sources. There are a total of 63 different themes, including hawker/food centres, hotels, monuments, museums, parks, supermarkets, and historic sites.

\begin{figure}[h]
\centering
\includegraphics[width=0.45\textwidth]{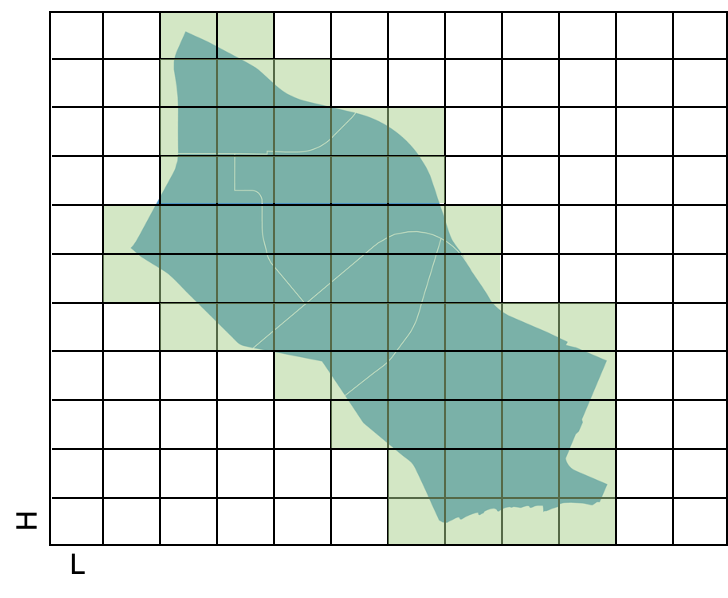}
\caption{The data procurement step begins by defining the dimensions of a rectangular bounding box that envelopes the study area before dividing the bounding box into a grid pattern consisting of sub-bounding boxes. The study area's shapefile is subsequently used to filter out all sub-bounding boxes that do not lie within its boundaries.\label{fig:dataprocurement}}
\end{figure}

On the other hand, Google Places and HERE Map require users to provide the geographic coordinates for the region of interest, formatted as a rectangular bounding box or bounding sphere. For these data sources, the data procurement step was performed by defining a rectangular bounding box that envelopes the entire study area before dividing the bounding box into a grid format consisting of sub-bounding boxes of size \textit{L} metres by \textit{H} metres. The study area's shapefile is subsequently used to filter out the sub-bounding boxes that do not lie within the study area's boundary to speed up the data procurement process. Figure \ref{fig:dataprocurement} provides a graphical representation of the steps described above.

\begin{figure}[h]
\centering
\includegraphics[width=0.5\textwidth]{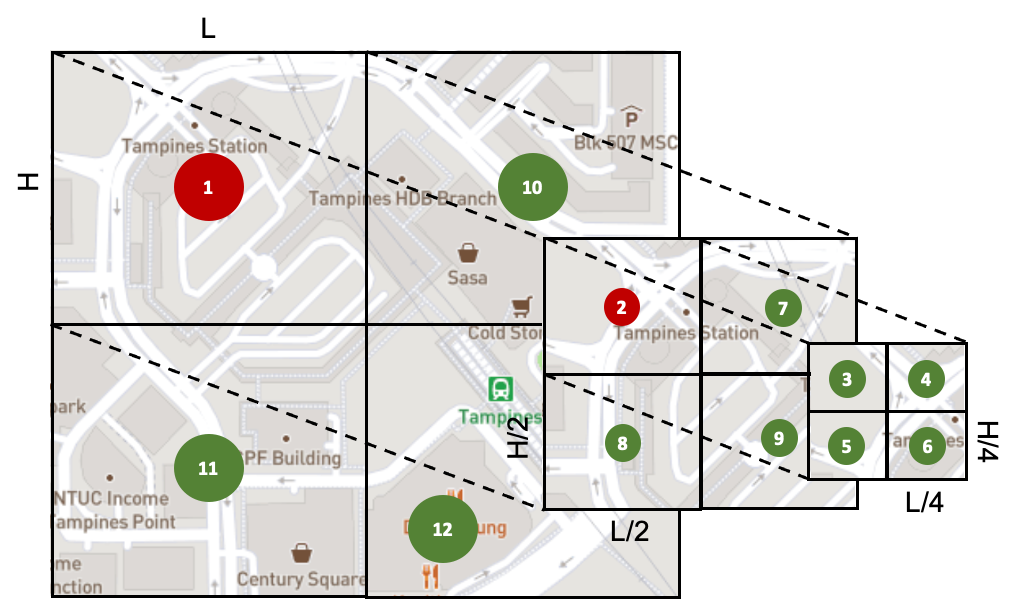}
\caption{The variable bounding box approach starts by dividing the study area into a series of sub-bounding boxes of a certain dimension and recursively dividing each sub-bounding box into boxes of smaller dimensions each time the number of results returned reaches the upper limit. The number in each circle represents the sequence in which an example query was constructed and called. The red circle indicates that the number of results returned reaches the upper limit, while a green circle indicates that the number of results returned falls below the upper limit. \label{fig:variableboundingbox}}
\end{figure}

Amongst the sub-bounding boxes that fall within the study area's boundary, their exact dimensions (i.e., \textit{L} and \textit{H}) are defined using a variable bounding box strategy, similar to \cite{juhasz2017catch}, which adjusts itself depending on the concentration of POIs found within a particular region. The approach is implemented by iterating through each sub-bounding box and constructing query calls based on its coordinates. The number of results returned per query is subsequently checked to determine if it reaches an upper limit. Google Places, for instance, has set the maximum number of results returned per query at 20 results, with the inclusion of a token that can return up to a total of 60 results \cite{googleplacetypes2021}. If the upper limit is reached, the sub-bounding box is further divided into four smaller sub-bounding boxes of half the original dimensions (i.e., \textit{L/2} and \textit{H/2}) before constructing a new set of query calls based on their coordinates. This recursive process will continue until the bounding box dimensions fall below a minimum threshold of 25 metres or when the number of returned results falls below the upper limit. This approach allows us to construct smaller sub-bounding boxes in regions with a higher concentration of POIs, while wider sub-bounding boxes will be used in less concentrated regions to minimise any information loss. Figure \ref{fig:variableboundingbox} provides a graphical representation of the variable bounding box approach described above.

Lastly, a data cleaning step was performed to remove any duplicated POIs based on their unique identifier. 

\subsubsection{Step 2: Schema standardisation}

After procuring POI data from the five data sources, the first challenge arises where it was observed that each data source uses a unique schema and different data formats when representing the attributes of their POIs. This issue poses a significant challenge downstream when we attempt to match the POIs from different data sources to identify overlaps, as the matching process is usually performed by measuring the similarity of their POI attributes. Therefore, we overcame this challenge by formatting each POI to follow an identical custom schema to standardise its attribute names and data storage format. The schema follows the GeoJSON format due to its prevalent use in representing geospatial data and can support a wide variety of geographic data structures, including Point, LineString, Polygon, MultiPoint, MultiLineString, and MultiPolygon \cite{geojson2020}.

Other than standardising the representation of the POI attributes, the POI's address information was also segmented into different components using the \textit{libpostal} library \cite{libpostal2020} and rearranged to follow the same address sequence (i.e., block number -> street name -> state -> country). The library uses statistical natural language processing (NLP) techniques to parse and normalise the addresses from different geographical locations to ensure consistency between different user inputs. This step is crucial as addresses often contain local conventions, abbreviations, and regional context, which is hard to account for when performing machine comparisons. Through the schema standardisation step, the complete set of attributes captured in each POI have been standardised to consist of its geographic coordinates, address information, location name, place type, data source, a unique identifier, date of data procurement, and an attribute indicating whether the POI requires further verification. The purpose of including this last attribute is explained in the next subsection on taxonomy mapping.

\begin{figure}
\centering
\includegraphics[width=0.45\textwidth]{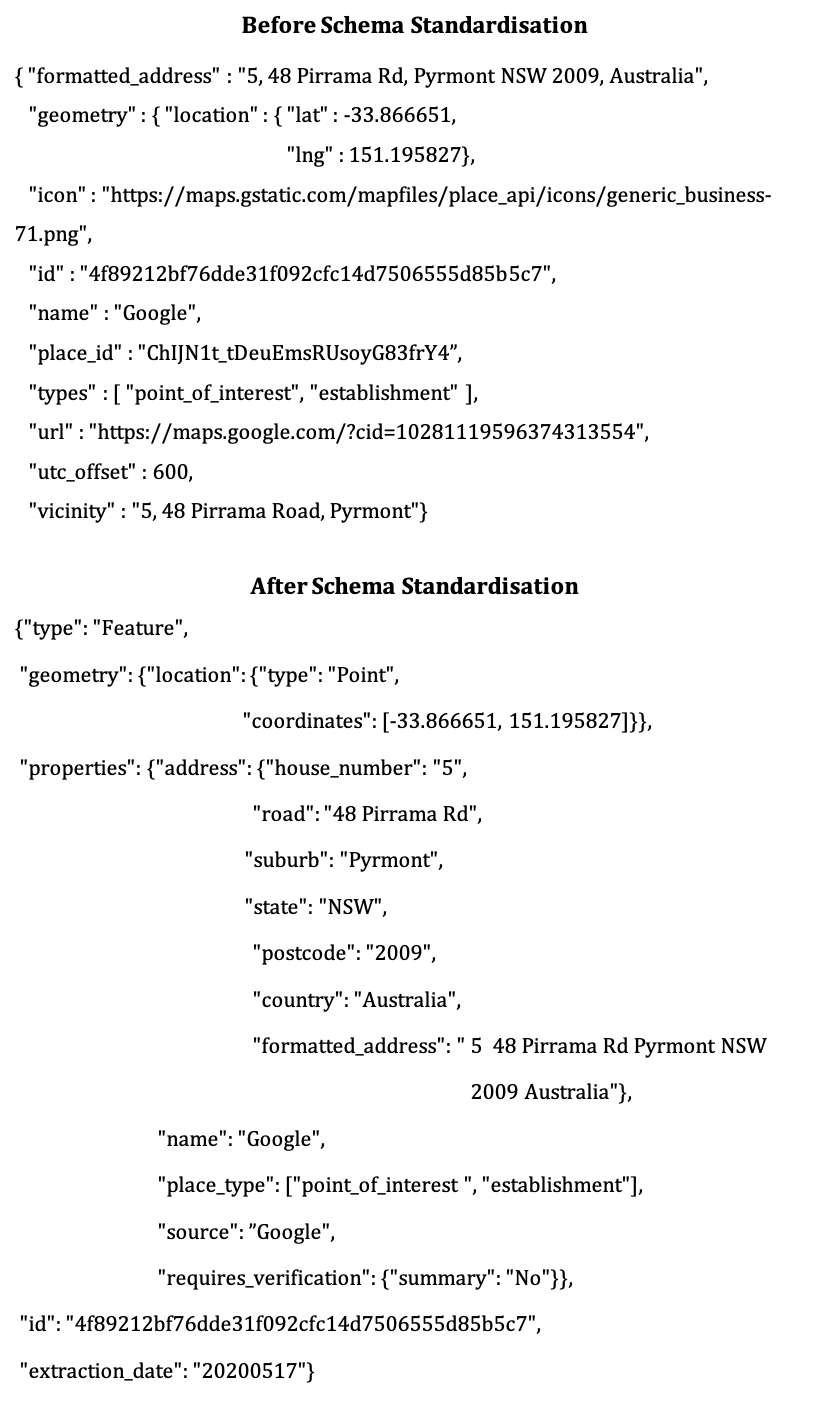}
\caption{An example POI from Google Places before and after the schema standardisation step. A dummy example is provided for illustration.\label{fig:schemastandardisation}}
\end{figure}

Figure \ref{fig:schemastandardisation} provides an example of a POI from Google Places before and after the schema standardisation step.

\subsubsection{Step 3: Taxonomy mapping}
After standardising the POI data procured to follow an identical custom schema, another challenge arises as different taxonomies were adopted by each data source when categorising their POI data. For instance, a "Restaurant" in Google Places can be categorised as an "Eating Establishment" in the SLA 2020 dataset, a "Hawker Centre" in OneMap, and a "Food Court" in OSM. Therefore, there is a need to overcome this issue by performing a taxonomy mapping step to ensure that all POIs follow a consistent place type taxonomy to aid the conflation process. That being said, the default place type taxonomy chosen for this study follows the taxonomy used in Google Places due to its comprehensive but non-overlapping coverage. However, users of the proposed framework can also adopt taxonomies from other data sources or create their custom taxonomies based on their unique needs.

The taxonomy mapping step is performed by first representing each place type as a mathematical word vector, where semantically similar words are placed close to each other in geometric space. This conversion of textual information to its mathematical representation is also known as word embedding. While many word embedding algorithms have been proposed by NLP researchers \cite{mikolov2013distributed, pennington2014glove} over the recent years, the fastText library was used in this study due to several key advantages. Unlike other word embedding algorithms that assign a distinct vector to each word, the model used within the fastText library is trained using a skip-gram method whereby each word in the training data is represented as a bag of character n-grams, and each character n-gram is associated with a vector representation \cite{joulin2016bag}. This allows the fastText model to represent each word as a sum of these vector representations, thereby allowing it to handle languages with large vocabularies, including rare words that did not appear in the training data \cite{bojanowski2017enriching}. For this study, the fastText model was pre-trained on 2 million word vectors with subword information from commoncrawl.org. The second advantage of using the fastText library to embed the place type information is due to its time efficiency, as it was able to report a similar classification performance compared to other deep learning classifiers while reporting a significantly shorter run time during model training and evaluation \cite{joulin2016bag}. 

After representing the POI's place type as a word vector $X$, it is compared against the word vectors from Google Places' taxonomy $Y_{google}$ by calculating their cosine similarity scores using Equation \ref{cosinesimilarity}. Given that the resulting similarity score ranges between 0 to 1, with a maximum score of 1 indicating that the two words are semantically identical, a high threshold value of 0.95 was chosen such that a mapping between the original place type and the new place type can only occur between semantically similar terms. In the case that the original place type cannot be mapped to any of the place types found within Google Places' taxonomy, the original place type will be retained, and this issue will be indicated in the \textit{requires\_verification} attribute so that it can be resolved in the data verification step. Furthermore, if the original place type contains multiple words such as "\textit{Asian Restaurant}", the entire phrase will be broken into its word components (i.e., "\textit{Asian}", "\textit{Restaurant}", and "\textit{Asian Restaurant}") before performing the same mapping step for each component. Therefore, a single place type can potentially be mapped to \textit{m} multiple place types under Google's taxonomy through this approach.

\begin{equation}
S_{cosine}(X,Y_{google}) = \frac{X \cdot Y_{google}}{||X||*||Y_{google}||}
\label{cosinesimilarity}
\end{equation}

\subsubsection{Step 4: POI matching}
The POI matching step is performed in two stages while considering three factors related to spatial similarity, name similarity, and address similarity.

In the first stage, the spatial similarity between each POI pair is considered by first filtering out all neighbouring POIs that fall within 100 metres of a centroid POI of interest. These neighbouring POIs are all treated equally as potential matches to the centroid POI as past studies \cite{li2020different, li2016entropy} have observed instances where matching POIs from different data sources can be found up to 100 metres apart due to human input error.

The second stage of the POI matching process is subsequently performed between each neighbouring POI and the centroid POI of interest by calculating their name and address similarity metrics. The name similarity metric is calculated by first tokenising the name information of each POI pair and sorting them based on alphabetical order before calculating the Levenshtein Distance between the two resulting strings. This process is implemented using the TokenSortRatio function in the \textit{Fuzzywuzzy} library \cite{fuzzywuzzy2020} before performing normalisation to result in a similarity score between 0 to 1 for each POI pair. While a string comparison approach may work well when comparing the names of two distinct locations, the same assumption does not hold when dealing with address information. Neighbouring POIs often have very similar address information that might only differ in terms of a few characters (i.e., street number or block number) but represent entirely different locations. Therefore, using a string comparison approach to calculate the address similarity metric is not appropriate as it places equal weight on each matching string between a pair of POIs. Instead, a weighted approach was adopted in this study by placing a heavier weight on matches for specific words that occur less frequently (i.e., block number, street number) while placing a smaller weight on frequently occurring words (i.e., street name, state, country) found in the addresses of neighbouring POIs. This weighted approach is achieved by applying the concept of Term Frequency-Inverse Document Frequency (TF-IDF) from statistical NLP \cite{qaiser2018text}. In information retrieval, TF-IDF is a numerical statistic that reflects the importance of a word relative to the document and other documents in the same collection. Based on Equation \ref{tfidf}, the TF-IDF statistic increases proportionally based on the number of times a word $t$ appears in document $d$ but is offset when the same word appears in multiple documents $D$. In this context, each document corresponds to the address of a neighbouring POI, while the collection of documents refers to the addresses of the neighbouring POIs. The address similarity metric between each POI pair is thus obtained by calculating the cosine similarity score (refer to \ref{cosinesimilarity}) between their address information vectorised using the TF-IDF statistic.

\begin{equation}
TF-IDF(t,d) = TF(t,d) * IDF(t,d)
\label{tfidf}
\end{equation}

\noindent
where

\begin{equation}
TF(t,d)=\frac{\sum_{j \in d} 1_{j=t}}{|d|}
\end{equation}

\begin{equation}
IDF(t,d)=\log{\frac{|D|}{\sum_{k \in D} 1_{t \in k}}}
\end{equation}

The resulting name similarity and address similarity metrics for each POI pair are subsequently passed into a binary ML classifier to determine if they match. The details of the classifier's implementation process are covered in Section \ref{sec:modelimplementation}. 

Due to the generalisability of this framework, the POI matching approach adopted in this study can also be replaced by other POI matching approaches discussed in Section \ref{sec:poimatching}. However, it should be noted that some of the approaches reviewed require the availability of specific attributes (i.e., user description, topic, website), which may not be captured by all data sources, thus limiting their viability.

\subsubsection{Step 5: POI unification}
Once the matching POIs are identified, they are merged in the POI unification step to form unique POIs following the merging rules listed in Table \ref{tab:mergingrules}. By ranking each matching POI based on their sources' reliability, the final geometric location is obtained by finding the centroid of the POIs from the most authoritative source, while the final address and location name are determined by selecting the longest address and name strings from the same group of trusted POIs. The POI sources used in this study are ranked from the most authoritative to the least authoritative in the following order: government agencies (i.e., OneMap followed by the SLA 2020 dataset), LBSNs (i.e., Google Places), commercial data providers (i.e., HERE Map), and open-source projects (i.e., OSM). The rest of the attributes are obtained by performing a union of their corresponding attributes from each matching POI to retain the maximum amount of information. Finally, any POIs that do not have any place type information after this unification step will be highlighted in the \textit{requires\_verification} attribute.

\begin{table}
\centering
\caption{Merging rules for POI matches.}
\label{tab:mergingrules}
\setlength{\tabcolsep}{3pt}
\begin{tabular}{ll}
\hline
Attributes &
  Merging Rules \\ \hline
Geometric location &
  \begin{tabular}[c]{@{}l@{}}Rank the POI matches based on the reliability \\ of their respective sources and calculate the \\ centroid of the most highly ranked POIs.\end{tabular} \\ \hline
Geometric bound &
  \begin{tabular}[c]{@{}l@{}}Select the most conservative (or largest)\\ bounds among all POI matches.\end{tabular} \\ \hline
Address, Name &
  \begin{tabular}[c]{@{}l@{}}Rank the POI matches based on the \\ reliability of their respective sources and \\ select the longest attribute string among\\ the most highly ranked POIs.\end{tabular} \\ \hline
\begin{tabular}[c]{@{}l@{}}Place type, Location, \\ Tags, Data source, \\ Unique identifier\end{tabular} &
  Union of all POI matches. \\ \hline
Extraction date &
  \begin{tabular}[c]{@{}l@{}}The latest extraction date among all matches.\end{tabular} \\ \hline
\textit{require\_verification} &
  \begin{tabular}[c]{@{}l@{}}If any of the POI matches require verification,\\ the unified POI will also require verification.\end{tabular} \\ \hline
\end{tabular}
\end{table}

\subsubsection{Step 6: Data verification}
In the final data verification step, POIs which require verification are identified and filtered out via the \textit{requires\_verification} attribute. Based on the previous steps described, there are two reasons why a POI would require verification. The first reason is due to the lack of an appropriate mapping between the POI's original place type and Google Places' place type taxonomy. This issue is resolved by performing these mappings through manual intervention. The second reason is due to the POIs lacking a place type category after the POI unification step. This scenario occurs when the original POI was initially missing its place type category, and it was unable to match with any of its neighbouring POIs with place type information. Since none of the POIs from all data sources have missing place type information, no POIs in the final unified dataset fell into this category.

\begin{figure}[t]
\centering
\includegraphics[width=0.75\textwidth]{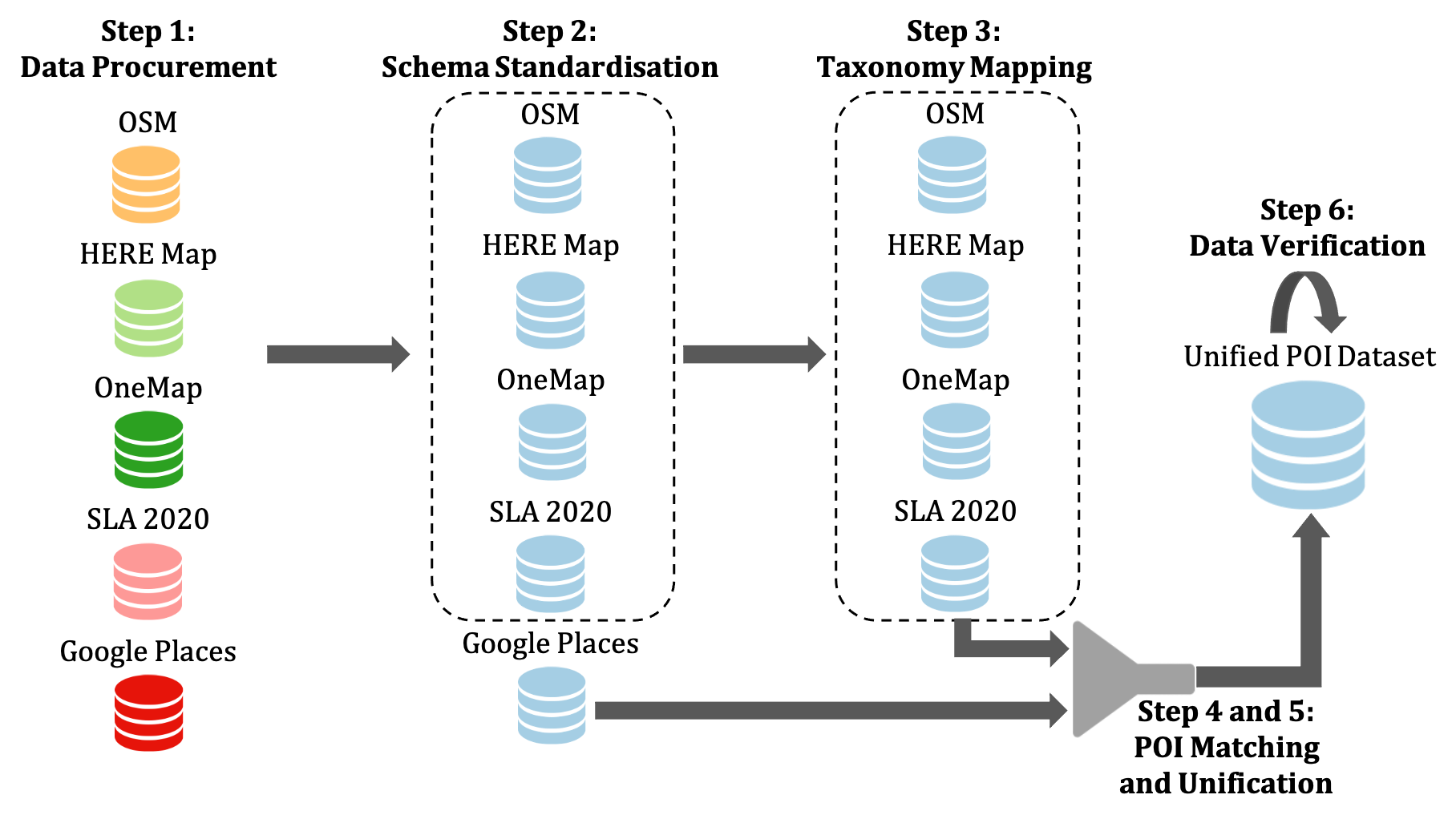}
\caption{Application of the proposed POI conflation framework within the study area of Tampines. POIs obtained from Google Places are not required to go through the Taxonomy Mapping step as the Google Places' place type taxonomy was chosen as the default taxonomy in this study. \label{fig:framework}}
\end{figure}

Figure \ref{fig:framework} provides a graphical representation of the proposed POI conflation framework applied within the study area involving the five data sources. The relevant source code is also made publicly available in an online code repository \cite{github2021}.

\section{Model implementation}
\label{sec:modelimplementation}
Given that the proposed POI matching model uses a supervised ML classifier to identify matches between each pair of POIs, the ground truth data used for training the ML classifier is obtained by procuring the POI data from another region located in the eastern part of Singapore (Southwest: 1.331747, 103.961258; Northwest: 1.339397, 103.961258; Northeast: 1.339397, 103.969027; Southeast: 1.331747, 103.969027) and manually labelling all POI matches and non-matches that occur between the five data sources. The region of interest spans over 0.75km\textsuperscript{2} and contains a business park where different technology companies, software enterprises, and research and development offices are situated. A retail mall and transit hub are also located in the vicinity, resulting in a diverse composition of POIs related to food, entertainment, transportation, and commercial activities. Based on the combination of the five data sources considered, a total of 1,227 POIs were found within the region with 200 pairs of POI matches (2.3\%) and 8,498 pairs of non-matches. 

Due to the significant imbalance between the number of POI matches and non-matches found in the labelled dataset, the development of an ML classifier for POI matching will be naturally biased towards the majority class (i.e., non-matches), potentially resulting in poorer model performance, especially when identifying POI matches. This class imbalance issue observed in the labelled dataset also reflects reality where it is significantly more likely to find POI non-matches than matches when comparing any neighbouring pair of POIs. 

To overcome this issue, we followed a similar approach proposed in a previous study \cite{low2020commercial} by using a combination of hybrid sampling techniques, bootstrap aggregation, and ensemble models to develop our POI matching classifier. The approach involves randomly splitting the labelled data into a training set and test set following a 75/25 ratio for both the minority class (i.e., POI matches) and the majority class (i.e., POI non-matches) separately. The training data for the minority class is subsequently randomly oversampled while we performed random undersampling on the majority class before combining these data samples to create multiple datasets containing an equal number of POI matches and non-matches. The final step involves training an ensemble classifier on each dataset and optimising each model through hyperparameter tuning using a 5-fold cross-validation approach. During model inference, the name and address similarity scores of each POI pair, involving a neighbouring POI and the centroid POI of interest, are passed separately into each of these models before combining their classification probabilities via averaging to obtain the most probable match result.

\section{Evaluation and discussion}
\label{sec:evaluation}
In this section, the proposed POI conflation framework is evaluated by comparing the unified POI dataset against the five POI data sources (i.e., OSM, Google Places, HERE Map, OneMap, and the SLA 2020 dataset) in terms of data coverage and completeness. Furthermore, the proposed POI matching approach is also evaluated against other baseline matching approaches based on its matching accuracy.

\subsection{Data coverage and completeness}

\begin{table*}[t]
\centering
\caption{The data coverage and completeness of each data source, including the unified dataset, broken down to the attribute level. A percentage value is provided to reflect the fraction of POIs that contains a particular attribute within each data source.}
\label{tab:datacoverage}
\setlength{\tabcolsep}{3pt}
\begin{tabular}{ccccccc}
\hline
\begin{tabular}[c]{@{}c@{}}Data\\ Source\end{tabular} &
  \begin{tabular}[c]{@{}c@{}}Geographic\\ Coordinates\end{tabular} &
  Address &
  Name &
  Place Type &
  Tags &
  Number of POIs \\ \hline
OSM             & 385 (100\%)   & 291 (75.6\%)   & 126 (32.7\%)   & 385 (100\%)   & 0 (0\%)        & 385   \\ \hline
Google Places   & 7,835 (100\%) & 7,425 (94.8\%) & 7,834 (99.9\%) & 7,835 (100\%) & 0 (0\%)        & 7,835 \\ \hline
HERE Map        & 2,187 (100\%) & 2,163 (98.9\%) & 2,187 (100\%)  & 2,187 (100\%) & 510 (23.3\%)   & 2,187 \\ \hline
OneMap          & 1,220 (100\%) & 846 (69.3\%)   & 1,220 (100\%)  & 1,220 (100\%) & 1,091 (89.4\%) & 1,220 \\ \hline
SLA Dataset     & 479 (100\%)   & 479 (100\%)    & 479 (100\%)    & 479 (100\%)   & 479 (100\%)    & 479   \\ \hline
\textbf{Unified Dataset} & \textbf{8,699 (100\%)} & \textbf{8,093 (93.0\%)} & \textbf{8,698 (99.9\%)} & \textbf{8,699 (100\%)} & \textbf{1,353 (15.6\%)} & \textbf{8,699} \\ \hline
\end{tabular}
\end{table*}

\begin{figure*}[t]
\centering
\includegraphics[width=0.75\textwidth]{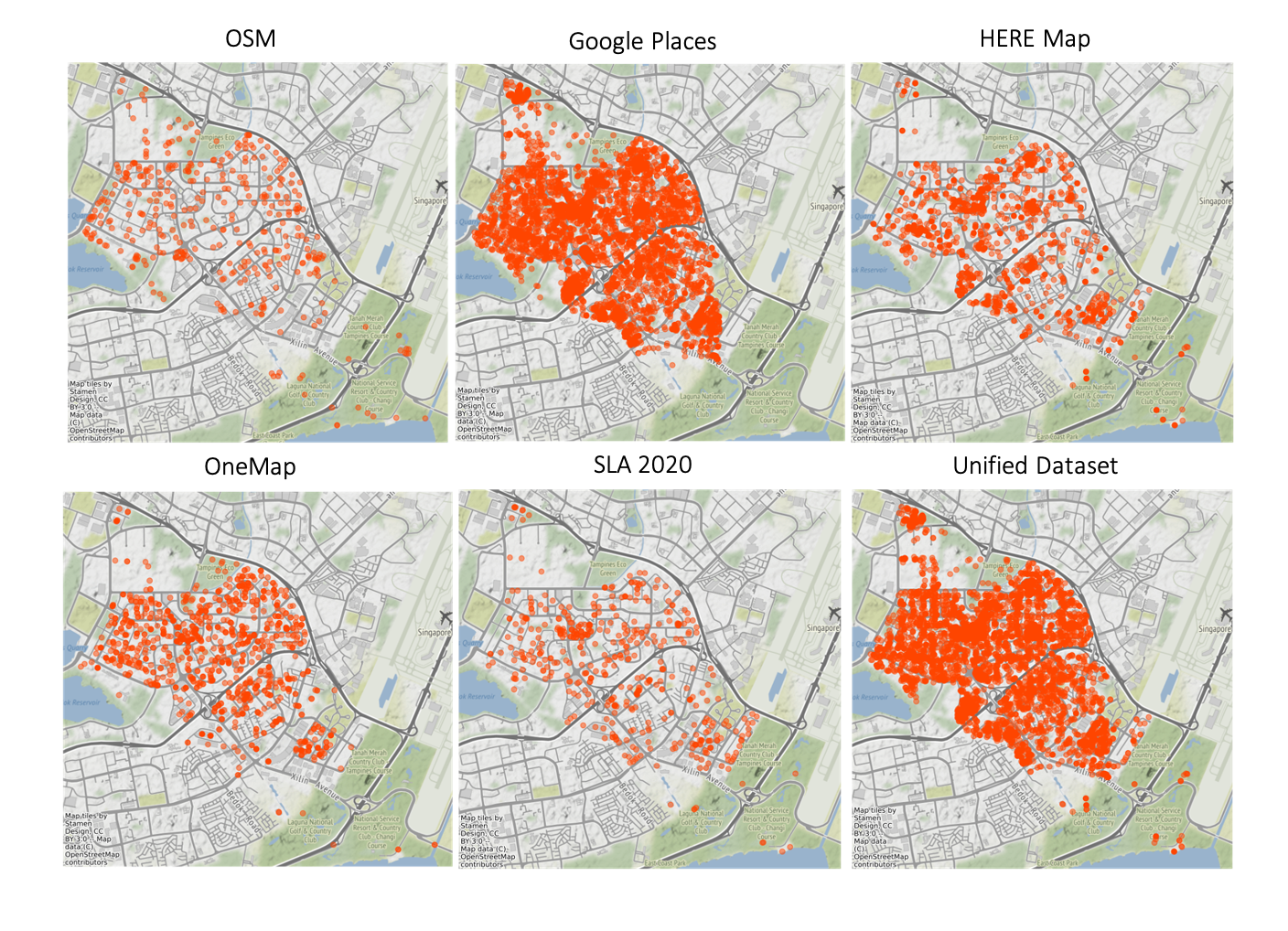}
\caption{The geographical distribution of the POIs from different data sources. \label{fig:poidistribution}}
\end{figure*}

Based on the POI data obtained from the five data sources within the study area, Table \ref{tab:datacoverage} reflects the data coverage and completeness of each data source, calculated down to the attribute level. These attributes include the geographic coordinates, address, location name, place type, tags, and number of POIs from each data source. It should be highlighted that the results reported in Table \ref{tab:datacoverage} are calculated before performing the data verification step. It can be observed that the data coverage of the unified dataset was more comprehensive compared to any of the five data sources considered in this study for all attributes. Furthermore, out of the 12,106 POIs that were procured from the five data sources, we were able to identify 3,407 POI matches (28.1\%) and performed data unification to end up with 8,699 unique POIs. This result indicates a significant overlap between the different data sources, and the proposed POI conflation framework was able to successfully process, identify, and merge the overlapping POIs to obtain a more comprehensive and complete POI dataset. Furthermore, given that the total run time for the POI matching and unification steps could be completed under 3 minutes, this further demonstrates the approach's scalability for large scale implementation in dense urban contexts. Figure \ref{fig:poidistribution} depicts the geographical distribution of the POIs from the five data sources, together with the POIs from the unified dataset.

\subsection{Matching accuracy}

\subsubsection{Evaluation metrics}
The matching accuracy of the proposed POI matching approach is evaluated based on overall accuracy and balanced accuracy. While overall accuracy is a standard evaluation metric used frequently in past studies \cite{mckenzie2014weighted, novack2018graph, almeida2018automatic}, the second evaluation metric (i.e., balanced accuracy) provides a more appropriate representation of the approach's performance by placing equal weights on the model's ability to identify both POI matches and non-matches during evaluation. This evaluation metric allows us to address the significant imbalance between the number of POI matches and non-matches usually found between POI datasets.

Overall accuracy is measured by calculating the cumulative true-positive $TP$, true-negative $TN$, false-positive $FP$, and false-negative $FN$ values before computing the fraction of true results against all instances, as shown in Equation \ref{overallaccuracy}.

\begin{equation}
accuracy_{overall} = \frac{TP+TN}{TP+TN+FP+FN}
\label{overallaccuracy}
\end{equation}

On the other hand, balanced accuracy involves computing an average of the same accuracy measure expressed in Equation \ref{overallaccuracy} for the majority and minority classes separately using $TP_{i}$, $TN_{i}$, $FP_{i}$, and $FN_{i}$, where $i\in\{match,non-match\}$, as shown in Equation \ref{balancedaccuracy}.

\begin{equation}
accuracy_{balanced} = \frac{\sum_{i}^{\{match,non-match\}}\frac{TP_{i}+TN_{i}}{TP_{i}+TN_{i}+FP_{i}+FN_{i}}}{|\{match, non-match\}|}
\label{balancedaccuracy}
\end{equation}

\subsubsection{Baselines}
Apart from evaluating the proposed POI matching approach based on the two evaluation metrics described above, it will also be evaluated against other baseline approaches described below.

\textit{String}: The first baseline matching approach uses a string comparison method to calculate the name and address similarity scores ($S_{name}$ and $S_{address}$) between each POI pair before combining the scores using a weighted sum aggregation approach to produce a final similarity score $S_{WSA}$ between 0 and 1. The optimal threshold value $V_{threshold}$ for determining POI matches and the coefficients for aggregating the name and address similarity scores ($\alpha$ and $\beta$) are determined by evaluating the performance of different coefficient combinations on a hold-out set.

\begin{equation}
f(S_{name}, S_{address}) = \begin{cases}
match, & \text{if } S_{WSA} > V_{threshold}\\ 
non-match, & \text{otherwise}
\end{cases}
\end{equation}

where
\begin{equation}
S_{WSA} = \alpha S_{name} + \beta S_{address}
\end{equation}

\begin{equation}
\alpha, \beta, S_{WSA}, S_{name}, S_{address}, V_{threshold} \in [0,1]
\end{equation}

\textit{TF-IDF}: The second baseline approach follows a similar idea as the first approach by replacing the string comparison method with TF-IDF. More specifically, the name and address strings of each POI pair are first vectorised using TF-IDF before calculating their similarity scores using the cosine similarity equation expressed in Equation \ref{cosinesimilarity}. The rest of the steps for identifying POI matches after calculating the similarity scores are identical to the first baseline approach.

\textit{String + TF-IDF}: The third baseline approach uses a hybrid combination of string comparison method for calculating the name similarity score and TF-IDF to calculate the address similarity score. Both scores are combined using a weighted sum aggregation approach to identify POI matches that exceed a specific threshold value.

\textit{String + ML}: The fourth baseline approach is an extension of \textit{String} by passing the name and address similarity scores as input features into an ML classifier to identify POI matches, instead of using a weighted sum aggregation approach.

\textit{TF-IDF + ML}: The fifth baseline approach is an extension of \textit{TF-IDF} by passing the name and address cosine similarity scores as input features into an ML classifier to identify POI matches, instead of using a weighted sum aggregation approach.

\textit{String + TF-IDF + ML}: The sixth baseline approach is a simplified version of the proposed POI matching approach by skipping the hybrid sampling and bootstrap aggregation steps to rebalance the majority and minority classes in the training dataset. 

\textit{String + ML + Data Rebalancing} and \textit{TF-IDF + ML + Data Rebalancing}: Lastly, the seventh and eighth baseline approaches are an extension of \textit{String + ML} and \textit{TF-IDF + ML} by applying hybrid sampling techniques and bootstrap aggregation to rebalance the majority and minority classes in the training dataset before training the ML classifier.

Therefore, based on the naming conventions assigned to the seven baseline approaches, our proposed POI matching approach is represented as \textit{String + TF-IDF + ML + Data Rebalancing}.

\subsubsection{Classification algorithms}
Several ML classification algorithms were also evaluated during this study when developing the POI matching model to compare their performances. 

The first classification algorithm considered for evaluation is the Gradient Boosting (GB) algorithm. This algorithm follows an iterative functional gradient descent approach to minimise its loss function $L(y_{j},\gamma)$ by iteratively introducing a base learner (i.e., a decision tree) in a forward stage-wise fashion \cite{friedman2001greedy}. The model begins by initialising a constant function $F_{0}(x)$ that is incrementally updated by defining a decision tree $h_{m}(x)$ that improves the current model's performance $F_{m-1}(x)$ in the steepest descent direction, as shown in the equations below. Due to its robust performance, the GB algorithm has also being applied in many other application areas \cite{tekler2019using, low2020predicting}. 

\begin{equation}
F_{0}(x)=argmin_{\gamma}\sum_{j=1}^{n}L(y_{j},\gamma)
\end{equation}

\begin{equation}
F_{m}(x)=F_{m-1}(x)+argmin_{h_{m}\in H}[\sum_{j=1}^{n}L(y_{j},F_{m-1}(x_{j})+h_{m}(x_{j}))]
\end{equation}

\begin{equation}
=F_{m-1}(x)-\alpha_{m}\sum_{j=1}^{n}\nabla_{F_{m-1}}L(y_{j},F_{m-1}(x_{j}))
\end{equation}

where 
\begin{equation}
\alpha_{m}=argmin_{\alpha}[\sum_{j=1}^{n}L(y_{j},F_{m-1}(x_{j})\\
-\alpha \nabla L(y_{j},F_{m-1}(x_{j}))]
\end{equation}

The Bagging algorithm is another classification algorithm that aggregates the model output produced by relatively uncorrelated base learners (i.e., decision trees) to produce an ensemble model that is more powerful than any individual learner. The correlation between each learner is minimised by training them on different subsets of the original dataset, sampled with replacement. This algorithm is a more simplistic variant of the Random Forest (RF) algorithm, which further reduces each base learner's correlation by randomising the set of input features considered when splitting each decision tree node \cite{breiman2001random}. However, due to the small number of input features considered (i.e., address and name similarity scores), both algorithms' performance is unlikely to differ significantly. In some instances, the bagging algorithm was even able to outperform the RF algorithm despite its more simplistic implementation \cite{tekler2020near}.

The final classification algorithm considered for evaluation is the Support Vector Machine (SVM), which differs from the above classification algorithms as it does not produce an ensemble model. Instead, the algorithm constructs a hyperplane in an $n$-dimensional space (where $n$ equals the number of input features considered) that maximises its distance from the data points belonging to each distinct class. Given the imbalance between the number of POI matches and non-matches found in the labelled dataset, the algorithm can account for this imbalance by increasing the penalty hyperparameter $C$ when misclassifying a minority instance. This step involves multiplying $C$ with weight $w_{i}$, which is inversely proportional to the class frequency $n_{i}$ \cite{chang2011libsvm}. Therefore, the new penalty score for each class $C_{i}$ is redefined below, where $s$ refers to the total sample size and $l$ refers to the number of classes.

\begin{equation}
C_{i} = Cw_{i}= C\frac{s}{ls_{i}}
\end{equation}

\subsubsection{POI matching results}
The matching accuracy of the proposed POI matching approach is evaluated and presented in Table \ref{tab:matchingaccuracy}, together with the performance of the other baseline approaches defined at the start of the section. 

It can be observed from Table \ref{tab:matchingaccuracy} that the baseline approaches that use a weighted sum aggregation (WSA) method (i.e., \textit{String}, \textit{TF-IDF}, and \textit{String + TF-IDF}) tend to report high overall accuracy scores but experienced a significant performance drop when it comes to balanced accuracy. This result is due to the approaches' inability to account for the imbalance between the number of POI matches and non-matches, resulting in the models being overly biased towards the majority class (i.e., non-matches). However, by replacing the WSA method with an ML classifier to identify the POI matches (i.e., \textit{String + ML}, \textit{TF-IDF + ML} and \textit{String + TF-IDF + ML}), this performance drop was reduced slightly as the introduction of an ML approach led to a marginal increase in balanced accuracy while the overall accuracy experienced an insignificant drop. This result can be attributed to the ML model's increased complexity, which introduces a non-linear solution to the POI matching problem compared to the linear solution produced using the WSA method.

Furthermore, by combining the use of hybrid sampling techniques and bootstrap aggregation to rebalance the class distribution in the training dataset, we observed further improvements in the models' balanced accuracy scores. This observation holds regardless of whether a string comparison, TF-IDF, or a hybrid approach was used to calculate the name and address similarity metrics (i.e., \textit{String + ML}, \textit{TF-IDF + ML}, and \textit{String + TF-IDF + ML}). In the end, our proposed approach was able to outperform all baseline approaches by reporting the highest balanced accuracy scores when using a GB or Bagging model while, at the same time, closing the gap between overall accuracy and balanced accuracy. Furthermore, it can be observed from Table \ref{tab:matchingaccuracy} that the performance of the SVM model with adjusted class penalties was insufficient to address the class imbalance issue encountered in the labelled dataset, collaborating with findings from previous studies \cite{low2020commercial}.

Another notable observation from Table \ref{tab:matchingaccuracy} shows that the baseline approaches that use the weighted sum aggregation method (i.e., \textit{String}, \textit{TF-IDF}, and \textit{String + TF-IDF}) tend to place a significantly higher weight on the name similarity metric (i.e., $\alpha$) as compared to the address similarity metric (i.e., $\beta$). This occurrence is likely due to the observation that there tends to be less variability in the naming conventions of location names than addresses, which may contain abbreviations and missing information. Therefore, close matches in location names can be treated as a more reliable indicator for identifying POI matches compared to matches in the address information. An alternative explanation for placing a lower emphasis on the address information could be due to the high concentration of establishments that can be found within a densely populated city like Singapore. This setting naturally results in neighbouring POIs having very similar addresses, which provides less information during POI matching.

\begin{table*}
\centering
\caption{POI matching accuracy of the proposed approach (i.e., \textit{String + TF-IDF + ML + Data Rebalancing}), compared against other baseline approaches. Approaches using a machine learning model to perform POI matching do not need to define weighted sum parameters as it will be learned by the model.}
\label{tab:matchingaccuracy}
\setlength{\tabcolsep}{3pt}
\begin{tabular}{cllc}
\hline
Approach &
  \multicolumn{1}{c}{Balanced accuracy} &
  \multicolumn{1}{c}{Overall accuracy} &
  Optimal weighted sum parameter \\ \hline
\textit{String} &
  \multicolumn{1}{c}{0.906} &
  \multicolumn{1}{c}{0.984} &
  \begin{tabular}[c]{@{}c@{}}$V_{threshold}$: 0.85\\ $\alpha$: 0.80\\ $\beta$: 0.20\end{tabular} \\ \hline
\textit{TF-IDF} &
  \multicolumn{1}{c}{0.925} &
  \multicolumn{1}{c}{0.983} &
  \begin{tabular}[c]{@{}c@{}}$V_{threshold}$: 0.45\\ $\alpha$: 0.90\\ $\beta$: 0.10\end{tabular} \\ \hline
\textit{String + TF-IDF} &
  \multicolumn{1}{c}{0.897} &
  \multicolumn{1}{c}{0.985} &
  \begin{tabular}[c]{@{}c@{}}$V_{threshold}$: 0.85\\ $\alpha$: 0.95\\ $\beta$: 0.05\end{tabular} \\ \hline
\textit{String + ML} &
  \begin{tabular}[c]{@{}l@{}}Gradient Boosting: 0.910\\ Bagging: 0.905\\ SVM: 0.784\end{tabular} &
  \begin{tabular}[c]{@{}l@{}}Gradient Boosting: 0.981\\ Bagging: 0.976\\ \textbf{SVM: 0.992}\end{tabular} &
  NA \\ \hline
\textit{TF-IDF + ML} &
  \begin{tabular}[c]{@{}l@{}}Gradient Boosting: 0.945\\ Bagging: 0.947\\ SVM: 0.656\end{tabular} &
  \begin{tabular}[c]{@{}l@{}}Gradient Boosting: 0.981\\ Bagging: 0.967\\ SVM: 0.988\end{tabular} &
  NA \\ \hline
\textit{String + TF-IDF + ML} &
  \begin{tabular}[c]{@{}l@{}}Gradient Boosting: 0.957\\ Bagging: 0.971\\ SVM: 0.802\end{tabular} &
  \begin{tabular}[c]{@{}l@{}}Gradient Boosting: 0.982\\ Bagging: 0.957\\ SVM: 0.991\end{tabular} &
  NA \\ \hline
\textit{\begin{tabular}[c]{@{}c@{}}String + ML +\\ Data Rebalancing\end{tabular}} &
  \begin{tabular}[c]{@{}l@{}}Gradient Boosting: 0.929\\ Bagging: 0.927\\ SVM: 0.797\end{tabular} &
  \begin{tabular}[c]{@{}l@{}}Gradient Boosting: 0.977\\ Bagging: 0.972\\ SVM: 0.986\end{tabular} &
  NA \\ \hline
\textit{\begin{tabular}[c]{@{}c@{}}TF-IDF + ML +\\ Data Rebalancing\end{tabular}} &
  \begin{tabular}[c]{@{}l@{}}Gradient Boosting: 0.959\\ Bagging: 0.953\\ SVM: 0.673\end{tabular} &
  \begin{tabular}[c]{@{}l@{}}Gradient Boosting: 0.977\\ Bagging: 0.965\\ SVM: 0.976\end{tabular} &
  NA \\ \hline
\textit{\begin{tabular}[c]{@{}c@{}}\textbf{String + TF-IDF + ML}\\ \textbf{+ Data Rebalancing} \\ \textbf{(Proposed)}\end{tabular}} &
  \begin{tabular}[c]{@{}l@{}}\textbf{Gradient Boosting: 0.976}\\ \textbf{Bagging: 0.976}\\ SVM: 0.812\end{tabular} &
  \begin{tabular}[c]{@{}l@{}}Gradient Boosting: 0.972\\ Bagging: 0.956\\ SVM: 0.982\end{tabular} &
  NA \\ \hline
\end{tabular}
\end{table*}

\section{Conclusion}
\label{sec:conclusion}
This study proposes a novel end-to-end POI conflation framework that consists of six steps, starting with data procurement, schema standardisation, taxonomy mapping, POI matching, POI unification, and data verification. The feasibility of the proposed framework was demonstrated in a case study conducted in the eastern region of Singapore, where the POI data from five data sources was conflated to form a unified POI dataset. Based on a thorough evaluation performed on the proposed framework, the resulting unified dataset's data coverage and completeness were more comprehensive than any of the five POI data sources considered for this study. Furthermore, the proposed POI matching approach was also able to outperform all baseline approaches with a matching accuracy of 97.6\% with an average run time below 3 minutes when matching over 12,000 POIs, thereby demonstrating the proposed approach's viability for large scale implementation in dense urban contexts.

Through the application of the proposed POI conflation framework, the availability of a more comprehensive and high-quality POI dataset will serve as a valuable source of data for many geospatial applications, especially in many transportation and urban planning studies. For instance, a richer dataset can enable more accurate calculations of different accessibility measures to various essential services and amenities, such as retail malls, transportation hubs, and restaurants, providing more valuable insights into the area's reliance on e-commerce and food delivery services.

\newpage
\bibliographystyle{unsrtnat}
\bibliography{references}  

\begin{thebibliography}{72}
\providecommand{\natexlab}[1]{#1}
\providecommand{\url}[1]{\texttt{#1}}
\expandafter\ifx\csname urlstyle\endcsname\relax
  \providecommand{\doi}[1]{doi: #1}\else
  \providecommand{\doi}{doi: \begingroup \urlstyle{rm}\Url}\fi

\bibitem[Miller and Shaw(2015)]{miller2015geographic}
Harvey~J Miller and Shih-Lung Shaw.
\newblock Geographic information systems for transportation in the 21st
  century.
\newblock \emph{Geography Compass}, 9\penalty0 (4):\penalty0 180--189, 2015.

\bibitem[Tekler et~al.(2020{\natexlab{a}})Tekler, Low, Gunay, Andersen, and
  Blessing]{tekler2020scalable}
Zeynep~Duygu Tekler, Raymond Low, Burak Gunay, Rune~Korsholm Andersen, and
  Lucienne Blessing.
\newblock A scalable bluetooth low energy approach to identify occupancy
  patterns and profiles in office spaces.
\newblock \emph{Building and Environment}, 171:\penalty0 106681,
  2020{\natexlab{a}}.

\bibitem[Guidotti et~al.(2014)Guidotti, Monreale, Rinzivillo, Pedreschi, and
  Giannotti]{guidotti2014retrieving}
Riccardo Guidotti, Anna Monreale, Salvatore Rinzivillo, Dino Pedreschi, and
  Fosca Giannotti.
\newblock Retrieving points of interest from human systematic movements.
\newblock In \emph{International Conference on Software Engineering and Formal
  Methods}, pages 294--308. Springer, 2014.

\bibitem[Vhaduri et~al.(2017)Vhaduri, Poellabauer, Striegel, Lizardo, and
  Hachen]{vhaduri2017discovering}
Sudip Vhaduri, Christian Poellabauer, Aaron Striegel, Omar Lizardo, and David
  Hachen.
\newblock Discovering places of interest using sensor data from smartphones and
  wearables.
\newblock In \emph{2017 IEEE SmartWorld, Ubiquitous Intelligence \& Computing,
  Advanced \& Trusted Computed, Scalable Computing \& Communications, Cloud \&
  Big Data Computing, Internet of People and Smart City Innovation
  (SmartWorld/SCALCOM/UIC/ATC/CBDCom/IOP/SCI)}, pages 1--8. IEEE, 2017.

\bibitem[Touya et~al.(2017)Touya, Antoniou, Olteanu-Raimond, and
  Van~Damme]{touya2017assessing}
Guillaume Touya, Vyron Antoniou, Ana-Maria Olteanu-Raimond, and Marie-Dominique
  Van~Damme.
\newblock Assessing crowdsourced poi quality: Combining methods based on
  reference data, history, and spatial relations.
\newblock \emph{ISPRS International Journal of Geo-Information}, 6\penalty0
  (3):\penalty0 80, 2017.

\bibitem[Gong et~al.(2016)Gong, Liu, Wu, and Liu]{gong2016inferring}
Li~Gong, Xi~Liu, Lun Wu, and Yu~Liu.
\newblock Inferring trip purposes and uncovering travel patterns from taxi
  trajectory data.
\newblock \emph{Cartography and Geographic Information Science}, 43\penalty0
  (2):\penalty0 103--114, 2016.

\bibitem[Liu et~al.(2020)Liu, Tian, Zhang, and Wan]{liu2020identification}
Xudong Liu, Yongzhong Tian, Xueqian Zhang, and Zuyi Wan.
\newblock Identification of urban functional regions in chengdu based on taxi
  trajectory time series data.
\newblock \emph{ISPRS International Journal of Geo-Information}, 9\penalty0
  (3):\penalty0 158, 2020.

\bibitem[Low et~al.(2020{\natexlab{a}})Low, Cheah, and You]{low2020commercial}
Raymond Low, Lynette Cheah, and Linlin You.
\newblock Commercial vehicle activity prediction with imbalanced class
  distribution using a hybrid sampling and gradient boosting approach.
\newblock \emph{IEEE Transactions on Intelligent Transportation Systems},
  2020{\natexlab{a}}.

\bibitem[Rodrigues et~al.(2013)Rodrigues, Alves, Polisciuc, Jiang, Ferreira,
  and Pereira]{rodrigues2013estimating}
Filipe Rodrigues, Ana Alves, Evgheni Polisciuc, Shan Jiang, Joseph Ferreira,
  and F~Pereira.
\newblock Estimating disaggregated employment size from points-of-interest and
  census data: From mining the web to model implementation and visualization.
\newblock \emph{International Journal on Advances in Intelligent Systems},
  6\penalty0 (1):\penalty0 41--52, 2013.

\bibitem[Tekler et~al.(2019{\natexlab{a}})Tekler, Low, Chung, Low, and
  Blessing]{tekler2019waste}
Zeynep~Duygu Tekler, Raymond Low, Si~Ying Chung, Jonathan Sze~Choong Low, and
  Lucienne Blessing.
\newblock A waste management behavioural framework of singapore’s food
  manufacturing industry using factor analysis.
\newblock \emph{Procedia CIRP}, 80:\penalty0 578--583, 2019{\natexlab{a}}.

\bibitem[Tekler et~al.(2019{\natexlab{b}})Tekler, Low, and
  Blessing]{tekler2019alternative}
Zeynep~Duygu Tekler, Raymond Low, and Lucienne Blessing.
\newblock An alternative approach to monitor occupancy using bluetooth low
  energy technology in an office environment.
\newblock In \emph{Journal of Physics: Conference Series}, volume 1343, page
  012116. IOP Publishing, 2019{\natexlab{b}}.

\bibitem[Farshad et~al.(2013)Farshad, Li, Marina, and
  Garcia]{farshad2013microscopic}
Arsham Farshad, Jiwei Li, Mahesh~K Marina, and Francisco~J Garcia.
\newblock A microscopic look at wifi fingerprinting for indoor mobile phone
  localization in diverse environments.
\newblock In \emph{International conference on indoor positioning and indoor
  navigation}, pages 1--10. IEEE, 2013.

\bibitem[swa()]{swarm2020}
Swarm by foursquare.
\newblock \url{https://www.swarmapp.com/}.
\newblock Accessed: 2020-05-19.

\bibitem[goo({\natexlab{a}})]{googlemap2020}
Google maps platform.
\newblock \url{https://cloud.google.com/maps-platform/}, {\natexlab{a}}.
\newblock Accessed: 2020-05-19.

\bibitem[osm({\natexlab{a}})]{osm2020}
Openstreetmap.
\newblock \url{https://www.openstreetmap.org/about}, {\natexlab{a}}.
\newblock Accessed: 2020-05-19.

\bibitem[geo({\natexlab{a}})]{geonames2020}
Geonames.
\newblock \url{https://www.geonames.org/}, {\natexlab{a}}.
\newblock Accessed: 2020-05-19.

\bibitem[dun()]{dunbradstreet2020}
Dun \& bradstreet.
\newblock \url{http://www.dnb.com.sg/}.
\newblock Accessed: 2020-05-19.

\bibitem[inf()]{infousa2020}
Infousa.
\newblock \url{https://www.infousa.com/}.
\newblock Accessed: 2020-05-19.

\bibitem[one({\natexlab{a}})]{onemap2020}
Onemap.
\newblock \url{https://docs.onemap.sg/}, {\natexlab{a}}.
\newblock Accessed: 2020-05-19.

\bibitem[goo({\natexlab{b}})]{googleplacetypes2021}
Place types.
\newblock
  \url{https://developers.google.com/maps/documentation/places/web-service/supported_types},
  {\natexlab{b}}.
\newblock Accessed: 2021-09-06.

\bibitem[her({\natexlab{a}})]{here2020}
Here map.
\newblock \url{https://developer.here.com/products/geocoding-and-search},
  {\natexlab{a}}.
\newblock Accessed: 2020-05-19.

\bibitem[fou()]{foursquare2020}
Foursquare places.
\newblock \url{https://enterprise.foursquare.com/products/places}.
\newblock Accessed: 2020-05-19.

\bibitem[yel()]{yelp2020}
Yelp fusion.
\newblock \url{https://www.yelp.com/fusion}.
\newblock Accessed: 2020-05-19.

\bibitem[bai()]{baidu2020}
Baidu map.
\newblock \url{https://lbsyun.baidu.com/}.
\newblock Accessed: 2020-05-19.

\bibitem[wei()]{weibo2020}
Weibo.
\newblock \url{https://open.weibo.com/wiki/API}.
\newblock Accessed: 2020-05-19.

\bibitem[fac()]{facebook2020}
Facebook places.
\newblock \url{https://developers.facebook.com/products/places/}.
\newblock Accessed: 2020-05-19.

\bibitem[yah()]{yahoo2020}
Yahoo! maps.
\newblock \url{https://developer.yahoo.com/maps/rest/V1/}.
\newblock Accessed: 2020-05-19.

\bibitem[tri()]{tripadvisor2020}
Trip advisor.
\newblock \url{https://developer-tripadvisor.com/content-api/}.
\newblock Accessed: 2020-05-19.

\bibitem[gao()]{gaode2020}
Gaode map.
\newblock \url{https://lbs.amap.com/}.
\newblock Accessed: 2020-05-19.

\bibitem[Santos et~al.(2018{\natexlab{a}})Santos, Murrieta-Flores, and
  Martins]{santos2018learning}
Rui Santos, Patricia Murrieta-Flores, and Bruno Martins.
\newblock Learning to combine multiple string similarity metrics for effective
  toponym matching.
\newblock \emph{International Journal of Digital Earth}, 11\penalty0
  (9):\penalty0 913--938, 2018{\natexlab{a}}.

\bibitem[Santos et~al.(2018{\natexlab{b}})Santos, Murrieta-Flores, Calado, and
  Martins]{santos2018toponym}
Rui Santos, Patricia Murrieta-Flores, P{\'a}vel Calado, and Bruno Martins.
\newblock Toponym matching through deep neural networks.
\newblock \emph{International Journal of Geographical Information Science},
  32\penalty0 (2):\penalty0 324--348, 2018{\natexlab{b}}.

\bibitem[K{\i}l{\i}n{\c{c}}(2016)]{kilincc2016accurate}
Deniz K{\i}l{\i}n{\c{c}}.
\newblock An accurate toponym-matching measure based on approximate string
  matching.
\newblock \emph{Journal of Information Science}, 42\penalty0 (2):\penalty0
  138--149, 2016.

\bibitem[McKenzie et~al.(2014)McKenzie, Janowicz, and
  Adams]{mckenzie2014weighted}
Grant McKenzie, Krzysztof Janowicz, and Benjamin Adams.
\newblock A weighted multi-attribute method for matching user-generated points
  of interest.
\newblock \emph{Cartography and Geographic Information Science}, 41\penalty0
  (2):\penalty0 125--137, 2014.

\bibitem[Li et~al.(2016)Li, Xing, Xia, and Huang]{li2016entropy}
Lin Li, Xiaoyu Xing, Hui Xia, and Xiaoying Huang.
\newblock Entropy-weighted instance matching between different sourcing points
  of interest.
\newblock \emph{Entropy}, 18\penalty0 (2):\penalty0 45, 2016.

\bibitem[Li et~al.(2020)Li, Liu, Dai, and Liu]{li2020different}
Chengming Li, Li~Liu, Zhaoxin Dai, and Xiaoli Liu.
\newblock Different sourcing point of interest matching method considering
  multiple constraints.
\newblock \emph{ISPRS International Journal of Geo-Information}, 9\penalty0
  (4):\penalty0 214, 2020.

\bibitem[Novack et~al.(2018)Novack, Peters, and Zipf]{novack2018graph}
Tessio Novack, Robin Peters, and Alexander Zipf.
\newblock Graph-based matching of points-of-interest from collaborative
  geo-datasets.
\newblock \emph{ISPRS International Journal of Geo-Information}, 7\penalty0
  (3):\penalty0 117, 2018.

\bibitem[Psaila and Toccu(2019)]{psaila2019fuzzy}
Giuseppe Psaila and Maurizio Toccu.
\newblock A fuzzy technique for on-line aggregation of pois from social media:
  Definition and comparison with off-line random-forest classifiers.
\newblock \emph{Information}, 10\penalty0 (12):\penalty0 388, 2019.

\bibitem[Yu et~al.(2018{\natexlab{a}})Yu, Qiu, Liu, Lu, and
  Wan]{yu2018holistic}
Li~Yu, Peiyuan Qiu, Xiliang Liu, Feng Lu, and Bo~Wan.
\newblock A holistic approach to aligning geospatial data with multidimensional
  similarity measuring.
\newblock \emph{International Journal of Digital Earth}, 11\penalty0
  (8):\penalty0 845--862, 2018{\natexlab{a}}.

\bibitem[Almeida et~al.(2018)Almeida, Alves, and Gomes]{almeida2018automatic}
Alexandre Almeida, Ana Alves, and Rui Gomes.
\newblock Automatic poi matching using an outlier detection based approach.
\newblock In \emph{International Symposium on Intelligent Data Analysis}, pages
  40--51. Springer, 2018.

\bibitem[Jiang et~al.(2015)Jiang, Alves, Rodrigues, Ferreira~Jr, and
  Pereira]{jiang2015mining}
Shan Jiang, Ana Alves, Filipe Rodrigues, Joseph Ferreira~Jr, and Francisco~C
  Pereira.
\newblock Mining point-of-interest data from social networks for urban land use
  classification and disaggregation.
\newblock \emph{Computers, Environment and Urban Systems}, 53:\penalty0 36--46,
  2015.

\bibitem[Cohen et~al.(2003)Cohen, Ravikumar, Fienberg,
  et~al.]{cohen2003comparison}
William~W Cohen, Pradeep Ravikumar, Stephen~E Fienberg, et~al.
\newblock A comparison of string distance metrics for name-matching tasks.
\newblock In \emph{IIWeb}, volume 2003, pages 73--78, 2003.

\bibitem[Yang and Zhang(2015)]{yang2015pattern}
Bisheng Yang and Yunfei Zhang.
\newblock Pattern-mining approach for conflating crowdsourcing road networks
  with pois.
\newblock \emph{International Journal of Geographical Information Science},
  29\penalty0 (5):\penalty0 786--805, 2015.

\bibitem[Yu et~al.(2018{\natexlab{b}})Yu, McMeekin, Arnold, and
  West]{yu2018semantic}
Feiyan Yu, David~A McMeekin, Lesley Arnold, and Geoff West.
\newblock Semantic web technologies automate geospatial data conflation:
  conflating points of interest data for emergency response services.
\newblock In \emph{LBS 2018: 14th International Conference on Location Based
  Services}, pages 111--131. Springer, 2018{\natexlab{b}}.

\bibitem[{data.gov.sg}(2021{\natexlab{a}})]{datagovsg_master_2021}
{data.gov.sg}.
\newblock Master {Plan} 2014 {Planning} {Area} {Boundary} ({No} {Sea}),
  2021{\natexlab{a}}.
\newblock URL
  \url{data.gov.sg/dataset/master-plan-2014-planning-area-boundary-no-sea}.

\bibitem[lan()]{landarea2018}
Land area and dwelling units by town.
\newblock
  \url{https://data.gov.sg/dataset/land-area-and-dwelling-units-by-town?resource_id=898d985a-0996-4efd-b2c2-7d9fab4138e9}.
\newblock Accessed: 2020-05-19.

\bibitem[{data.gov.sg}(2021{\natexlab{b}})]{datagovsg2021}
{data.gov.sg}.
\newblock About us, 2021{\natexlab{b}}.
\newblock URL \url{https://data.gov.sg/about}.

\bibitem[osm({\natexlab{b}})]{osmoverpass2020}
Osm overpass api.
\newblock \url{https://wiki.openstreetmap.org/wiki/Overpass\_API},
  {\natexlab{b}}.
\newblock Accessed: 2020-05-19.

\bibitem[osm({\natexlab{c}})]{osmplanet2020}
Planet osm.
\newblock \url{https://wiki.openstreetmap.org/wiki/Planet.osm}, {\natexlab{c}}.
\newblock Accessed: 2020-05-19.

\bibitem[Haklay(2010)]{haklay_how_2010-1}
Mordechai Haklay.
\newblock How {Good} is {Volunteered} {Geographical} {Information}? {A}
  {Comparative} {Study} of {OpenStreetMap} and {Ordnance} {Survey} {Datasets}.
\newblock \emph{Environment and Planning B: Planning and Design}, 37\penalty0
  (4):\penalty0 682--703, August 2010.
\newblock ISSN 0265-8135.
\newblock \doi{10.1068/b35097}.
\newblock URL \url{https://doi.org/10.1068/b35097}.
\newblock Publisher: SAGE Publications Ltd STM.

\bibitem[goo({\natexlab{c}})]{google1012020}
Google maps 101: How we map the world.
\newblock
  \url{https://www.blog.google/products/maps/google-maps-101-how-we-map-world/},
  {\natexlab{c}}.
\newblock Accessed: 2020-05-19.

\bibitem[her({\natexlab{b}})]{heredata2020}
Here map data.
\newblock \url{https://www.here.com/products/mapping/map-data}, {\natexlab{b}}.
\newblock Accessed: 2020-05-19.

\bibitem[her({\natexlab{c}})]{hereapi2020}
Here map rest apis.
\newblock \url{https://developer.here.com/develop/rest-apis}, {\natexlab{c}}.
\newblock Accessed: 2020-05-19.

\bibitem[her({\natexlab{d}})]{herefeedback2020}
Here map submit feedback.
\newblock
  \url{https://developer.here.com/documentation/map-feedback/dev_guide/topics/quick-start-submit-feedback.html},
  {\natexlab{d}}.
\newblock Accessed: 2020-05-19.

\bibitem[one({\natexlab{b}})]{onemapthemes2021}
Themes.
\newblock \url{https://www.onemap.gov.sg/docs/#themes}, {\natexlab{b}}.
\newblock Accessed: 2021-09-06.

\bibitem[osm({\natexlab{d}})]{osmmapfeatures2021}
Map features.
\newblock \url{https://wiki.openstreetmap.org/wiki/Map_features},
  {\natexlab{d}}.
\newblock Accessed: 2021-09-06.

\bibitem[her({\natexlab{e}})]{hereplacetypes2021}
Place types.
\newblock
  \url{https://developer.here.com/documentation/map-feedback/dev_guide/topics/resource-type-place-type.html},
  {\natexlab{e}}.
\newblock Accessed: 2021-09-06.

\bibitem[Juh{\'a}sz and Hochmair(2017)]{juhasz2017catch}
Levente Juh{\'a}sz and Hartwig~H Hochmair.
\newblock Where to catch ‘em all?--a geographic analysis of pok{\'e}mon go
  locations.
\newblock \emph{Geo-spatial information science}, 20\penalty0 (3):\penalty0
  241--251, 2017.

\bibitem[geo({\natexlab{b}})]{geojson2020}
Geojson.
\newblock \url{https://geojson.org/}, {\natexlab{b}}.
\newblock Accessed: 2020-05-19.

\bibitem[lib()]{libpostal2020}
Libpostal.
\newblock \url{https://github.com/openvenues/libpostal}.
\newblock Accessed: 2020-05-19.

\bibitem[Mikolov et~al.(2013)Mikolov, Sutskever, Chen, Corrado, and
  Dean]{mikolov2013distributed}
Tomas Mikolov, Ilya Sutskever, Kai Chen, Greg~S Corrado, and Jeff Dean.
\newblock Distributed representations of words and phrases and their
  compositionality.
\newblock In \emph{Advances in neural information processing systems}, pages
  3111--3119, 2013.

\bibitem[Pennington et~al.(2014)Pennington, Socher, and
  Manning]{pennington2014glove}
Jeffrey Pennington, Richard Socher, and Christopher~D Manning.
\newblock Glove: Global vectors for word representation.
\newblock In \emph{Proceedings of the 2014 conference on empirical methods in
  natural language processing (EMNLP)}, pages 1532--1543, 2014.

\bibitem[Joulin et~al.(2016)Joulin, Grave, Bojanowski, and
  Mikolov]{joulin2016bag}
Armand Joulin, Edouard Grave, Piotr Bojanowski, and Tomas Mikolov.
\newblock Bag of tricks for efficient text classification.
\newblock \emph{arXiv preprint arXiv:1607.01759}, 2016.

\bibitem[Bojanowski et~al.(2017)Bojanowski, Grave, Joulin, and
  Mikolov]{bojanowski2017enriching}
Piotr Bojanowski, Edouard Grave, Armand Joulin, and Tomas Mikolov.
\newblock Enriching word vectors with subword information.
\newblock \emph{Transactions of the Association for Computational Linguistics},
  5:\penalty0 135--146, 2017.

\bibitem[fuz()]{fuzzywuzzy2020}
Fuzzywuzzy.
\newblock \url{https://github.com/seatgeek/fuzzywuzzy}.
\newblock Accessed: 2020-05-19.

\bibitem[Qaiser and Ali(2018)]{qaiser2018text}
Shahzad Qaiser and Ramsha Ali.
\newblock Text mining: use of tf-idf to examine the relevance of words to
  documents.
\newblock \emph{International Journal of Computer Applications}, 181\penalty0
  (1):\penalty0 25--29, 2018.

\bibitem[git()]{github2021}
Source code for poi conflation framework.
\newblock \url{https://github.com/iamraymondlow/poi-conflation-framework}.

\bibitem[Friedman(2001)]{friedman2001greedy}
Jerome~H Friedman.
\newblock Greedy function approximation: a gradient boosting machine.
\newblock \emph{Annals of statistics}, pages 1189--1232, 2001.

\bibitem[Tekler et~al.(2019{\natexlab{c}})Tekler, Low, and
  Blessing]{tekler2019using}
ZD~Tekler, R~Low, and L~Blessing.
\newblock Using smart technologies to identify occupancy and plug-in appliance
  interaction patterns in an office environment.
\newblock In \emph{IOP Conference Series: Materials Science and Engineering},
  volume 609, page 062010. IOP Publishing, 2019{\natexlab{c}}.

\bibitem[Low et~al.(2020{\natexlab{b}})Low, Tekler, and
  Cheah]{low2020predicting}
Raymond Low, Zeynep~Duygu Tekler, and Lynette Cheah.
\newblock Predicting commercial vehicle parking duration using generative
  adversarial multiple imputation networks.
\newblock \emph{Transportation Research Record}, page 0361198120932166,
  2020{\natexlab{b}}.

\bibitem[Breiman(2001)]{breiman2001random}
Leo Breiman.
\newblock Random forests.
\newblock \emph{Machine learning}, 45\penalty0 (1):\penalty0 5--32, 2001.

\bibitem[Tekler et~al.(2020{\natexlab{b}})Tekler, Low, Zhou, Yuen, Blessing,
  and Spanos]{tekler2020near}
Zeynep~Duygu Tekler, Raymond Low, Yuren Zhou, Chau Yuen, Lucienne Blessing, and
  Costas Spanos.
\newblock Near-real-time plug load identification using low-frequency power
  data in office spaces: Experiments and applications.
\newblock \emph{Applied Energy}, 275:\penalty0 115391, 2020{\natexlab{b}}.

\bibitem[Chang and Lin(2011)]{chang2011libsvm}
Chih-Chung Chang and Chih-Jen Lin.
\newblock Libsvm: a library for support vector machines.
\newblock \emph{ACM transactions on intelligent systems and technology (TIST)},
  2\penalty0 (3):\penalty0 27, 2011.

\end{thebibliography}






\end{document}